\documentclass[preprint,12pt]{elsarticle}

\usepackage{amssymb}
\usepackage{amsthm}

\usepackage{lineno}
\usepackage{amsmath}
\usepackage[ruled,vlined]{algorithm2e}
\usepackage{url}

\usepackage{color}
\newcommand{\bm}[1]{\mbox{\boldmath$#1$}}

\journal{arXiv}

\begin{document}

\begin{frontmatter}



\title{GlyphGAN: Style-Consistent Font Generation Based on Generative Adversarial Networks}


\author[Kyushu]{Hideaki Hayashi\corref{cor1}}
\address[Kyushu]{Department of Advanced Information Technology, Kyushu University, Fukuoka, Japan}
\cortext[cor1]{Corresponding author}
\ead{hayashi@ait.kyushu-u.ac.jp}
\author[Kyushu]{Kohtaro Abe}
\author[Kyushu]{Seiichi Uchida}

\begin{abstract}
In this paper, we propose GlyphGAN: 
style-consistent font generation based on generative adversarial networks (GANs). 
GANs are a framework for learning a generative model 
using a system of two neural networks competing with each other. 
One network generates synthetic images from random input vectors, 
and the other discriminates between synthetic and real images. 
The motivation of this study is to create new fonts using the GAN framework 
while maintaining style consistency over all characters. 
In GlyphGAN, the input vector for the generator network consists of two vectors: 
character class vector and style vector. 
The former is a one-hot vector and is associated with the character class of each sample image during training. 
The latter is a uniform random vector without supervised information. 
In this way, GlyphGAN can generate an infinite variety of fonts 
with the character and style independently controlled. 
Experimental results showed that fonts generated by GlyphGAN have style consistency 
and diversity different from the training images without losing their legibility.
\end{abstract}

\begin{keyword}
Font generation \sep generative adversarial networks \sep style consistency \sep deep convolutional neural network


\end{keyword}

\end{frontmatter}


 {\allowdisplaybreaks
	\section{Introduction}
	There is a variety of fonts in the world. 
As shown in Fig. \ref{01_various_fonts}, 
fonts are characterized by various components such as the thickness of lines, decoration, and serifs. 
There are also handwritten-like fonts, fonts made of outlines, fonts with lowercase letters capitalized, and so on. 
Among these fonts, the best ones will be chosen according to the medium such as books, newspapers, signboards, and web pages. 
Even for the same medium, different fonts can be used depending on the title, text, and speakers. 
In response to these demands, a large number of fonts have been created.
\begin{figure}[t]
		\centering
		\includegraphics[width=\hsize]{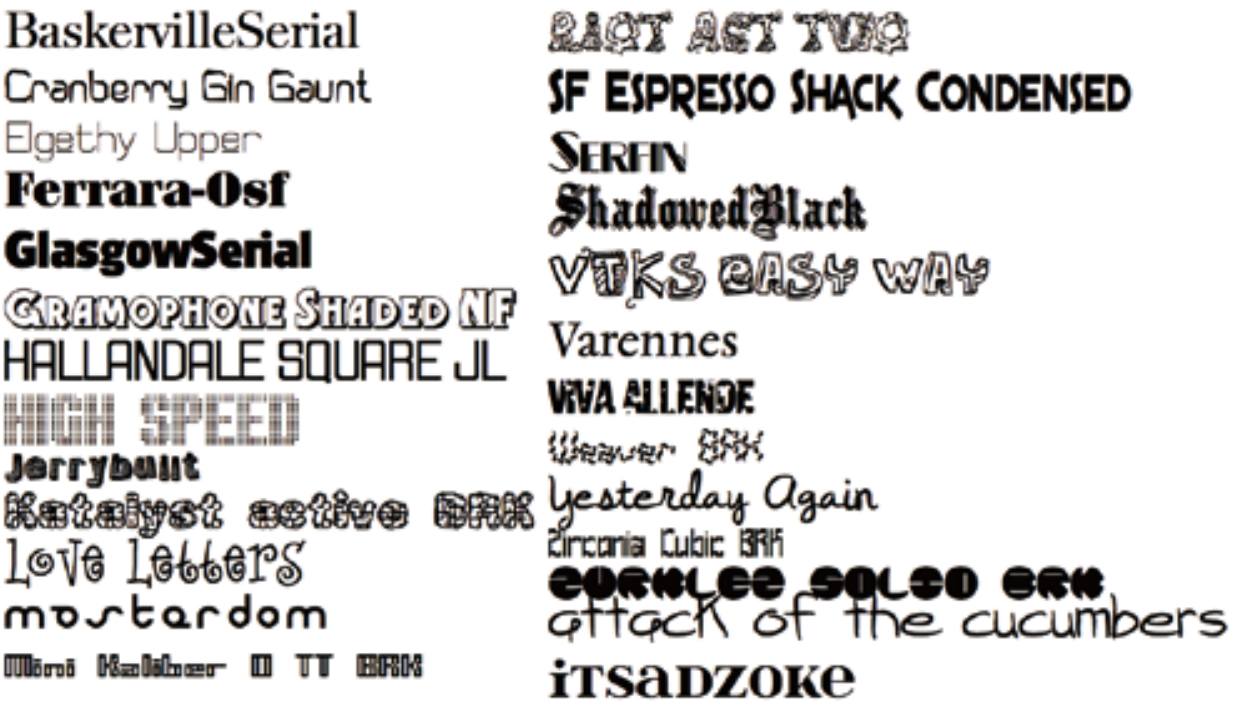}
	\caption{Examples of various fonts. Texts denote the name of the font.}
	\label{01_various_fonts}
\end{figure}

This study aims at the automatic design of fonts; 
a computer automatically generates various fonts instead of a human designing fonts individually.
There are two reasons why we aim at automatic design even though a large number of fonts already exists.

The first aim is to reduce the labor for creating a new font. 
Even today, new fonts are still being created. 
When a font is created, a large number of characters with the same style should be designed. 
In the case of alphabetic fonts, not only 52 upper/lowercase letters but symbols also are designed. 
For Japanese fonts, the labor increases because the Japanese language has a large number of letters including \textit{hiragana}, \textit{katakana}, and \textit{kanji}. 
Therefore, automatic font design can potentially reduce this labor by a large extent. 

The second aim is to understand designers' tacit knowledge via a constructive approach. 
Basically, fonts are created by designers individually. 
This know-how is fundamentally cultivated by the designer's experience and is not easily systematized. 
Reproduction of the process where a designer becomes able to create new fonts 
will lead to new knowledge of character design.

To realize the above aims, the following approaches can be considered. 
\begin{enumerate}
 	\item Designing all characters from a few samples: After manually designing a few examples as templates, the system automatically designs the remaining characters using these templates. This approach is effective, particularly for character sets with many of the same parts such as Chinese characters.
 	\item Transformation and interpolation: A new font is made of existing fonts via operations such as changing the thickness of lines, adding decorations, and calculating an interpolation of two fonts.
 	This approach has difficulty in designing a completely novel font 
 	because the generated font depends on the original font. 
 	\item Generating fonts automatically using machine learning: Utilizing a large number of fonts, a computer is trained to learn the design principle. If the computer can learn the designer's know-how, which is difficult to describe explicitly, then automatic font design with a high degree of freedom is realized. \label{01_by_ml}
\end{enumerate}
Studies related to the above approaches are described in the next section. 

This study focuses on approach \#\ref{01_by_ml}, i.e. machine learning-based font generation. This approach includes mainly two methods: transformation-based and generative model-based methods. In the former, a font is generated by adding style information to the existing font \cite{Atarsaikhan2017, Chang2017, Kaonashi2017, Lyu2017}. 
The latter estimates the manifold that the existing fonts compose in the image space, and then generates new font images by sampling data from the estimated manifold \cite{Bernhardsson2016blog, Bernhardsson2016git}. 
The latter has the potential to generate more diverse fonts although there are challenges in the manifold estimation and generation stability.

Related to the generative model-based method, generative adversarial networks (GANs) \cite{Goodfellow2014} 
have attracted much attention in terms of image generation. 
GANs are a framework for learning a generative model 
using a system of two neural networks competing with each other. 
One network generates synthetic images from a random input, 
and the other discriminates between synthetic and real images,  
thereby allowing the generation of highly realistic images. 
However, it is basically difficult to control the characteristics of the generated images using GANs 
because GANs generate images from random input. 
Considering the application to font generation, 
the generated font should have the same style for all characters. 

In this paper, we propose a font generation method based on GANs, which is named GlyphGAN. 
In GlyphGAN, the input vector for the generator network consists of two vectors: 
character class vector and style vector. 
The former is a one-hot vector and is associated with the character class of each sample during training. 
The latter is a uniform random vector without supervised information. 
In this way, GlyphGAN can generate an infinite variety of fonts 
with the character and style independently controlled. 

The main features of the proposed GlyphGAN are as follows: 
\begin{itemize}
	\item \textbf{Style consistency:} GlyphGAN can generate a font that has the same style over all characters. 
	\item \textbf{Legibility:} The generated fonts are legible compared with the other methods. 
	\item \textbf{Diversity:} The generated fonts have diversity different from the training images.
\end{itemize}

	\section{Related Work}\label{section:related}
\subsection{Example-based Font Generation}
Various attempts have been made in previous studies on automatic font design. 
One of the classical methods is example-based font generation \cite{Devroye1995,Tenenbaum2000,Suveeranont2009,Lake2015,Miyazaki2017,Yang2017}. 
For example, ``A'' is generated from human-designed ``B,'' ``C,'' $\ldots$, ``Z.''
Devroye and McDougall \cite{Devroye1995} proposed a method for creating a random printed handwriting font by perturbating a small sample set from a person's handwriting. 
Tenenbaum and Freeman \cite{Tenenbaum2000} used several font sets containing all alphabets to separate them into standard shapes of individual alphabets and font styles. 
Then, the style of example patterns was extracted by using the standard shape and applied to the other alphabets.
Suveeranont and Igarashi \cite{Suveeranont2009} proposed a model for generating a new font from a user-defined example. 
Miyazaki {\it et al.} \cite{Miyazaki2017} proposed an automatic typographic font generation method based on the extraction of strokes from a subset of characters. Yang {\it et al.} \cite{Yang2017} proposed a patch-based method to transfer heavy decoration from an example image to others.

\subsection{Transformation and Interpolation} 
Some studies attempted font generation based on the transformation and interpolation of existing fonts.
In \cite{Wada2006}, a transformation-based method was proposed in which new fonts are created by adjusting parameters such as the thickness, roundness, and slope of the font to reflect the sensibility input by a user.  
Wang {\it et al.} \cite{Wang2008} employed stroke-level transformation for generating Chinese letters. 
A new font can also be generated by interpolating multiple fonts \cite{Campbell2014,Uchida2015}. 
Campbell and Kautz \cite{Campbell2014} obtained a manifold of fonts by learning nonlinear mapping that can be used to smoothly interpolate between existing fonts. 
Uchida {\it et al.} \cite{Uchida2015} analyzed the distribution of fonts using a large-scale relative neighborhood graph, and then generated new fonts by using a contour-based interpolation between neighboring fonts. 

\subsection{Machine Learning}
Nowadays, various machine learning techniques are used for generating fonts. 
A recent trend is GAN-based methods, which will be reviewed subsequently.
Neural font style transfer \cite{Atarsaikhan2017} is an example of font generation using deep learning. 
This method transfers the style of one image to another input image using the features extracted from the intermediate layers of a convolutional neural network (CNN), 
inspired by the idea of neural style transfer \cite{Gatys2016}. 
In the neural font style transfer, various types of images are used as a style image such as textures and fonts of different languages from the input, 
thus expanding the possibility of font design by deep learning.

Other machine learning techniques have also been used for font or character pattern generation.
Lake {\it et al.} \cite{Lake2015} proposed an interesting way to generate handwritten patterns by Bayesian program learning. 
This approach infers a rule to draw an example pattern and then applies the rule to generate new patterns. 
Baluja \cite{Baluja2017} used a CNN-like neural network that is originally trained for font type discrimination. 
The neural network outputs a single letter or all alphabet letters from a limited number of examples.

\subsection{Font Generation by GANs}\label{section:FGGANs}
Recently, there are several attempts that utilize GANs for font generation.
The zi2zi \cite{Kaonashi2017} is a method that converts a certain font pattern into a target font pattern 
based on a combination of the pix2pix \cite{Isola2016}, AC-GAN \cite{Odena2017}, 
and domain transfer network \cite{Taigman2017}.
Although the generated fonts have sharp outlines and include a variety of styles,
the target font is restricted to those having a large number of character types such as Japanese, Chinese, and Korean; 
thus it is difficult to apply this method to alphabets that consist of few letters.

Many other methods have also been proposed. 
Chang and Gu \cite{Chang2017} proposed an example-based font generation by GANs that uses a U-net as its generator for character patterns with the target style. 
They claimed that their method is easier in balancing the loss functions than zi2zi.
Lyu {\it et al.} \cite{Lyu2017} used a GAN along with a supervised network, which is an autoencoder that captures the style of a target calligrapher.  
Azadi {\it et al.} \cite{Azadi2018} proposed an example-based font generation method by using a conditional GAN that is extended to deal with fewer examples. 
Lin {\it et al.} \cite{Lin2018} proposed a stroke-based font generation method where two trained styles can be interpolated by controlling a weight. 
Guo {\it et al.} \cite{Guo2018} used a skeleton vector of the target character and a font style vector (called a shape vector) as inputs for their GAN-based font generation network. 
Inspired by \cite{Campbell2014}, they also built a font manifold of those vectors and use it for generating various new font styles. 
Bhunia {\it et al.} \cite{Bhunia2018} used long short-term memory (LSTM) units in their generator to have a variable-length word image in a specific font. 

The main difference between the above GAN-based font generation methods and the proposed GlyphGAN 
lies in the way of providing input. 
The above methods are based on image-to-image transformation, 
where a new font is generated by adding style information to character class information 
extracted from a reference character image given as an input. 
This approach allows for a large number of character classes, 
whereas the generated font potentially depends on the shape of the input image; 
hence the generation of a completely novel font is difficult.  
Different from such an approach, 
GlyphGAN employs only abstracted inputs as vectors, 
thereby allowing the generation of fonts not seen in the training image. 
Although it is difficult to maintain legibility and style consistency in this approach, 
we manage to improve these important natures by embedding both the character ID and style information into the latent vector and introducing the loss function of the Wasserstein-GAN gradient penalty.

	\section{Preliminary Knowledge of Generative Adversarial Networks}
\subsection{Overview}
GANs are a framework for estimating a generative model composed of two neural networks called the generator $G$ and the discriminator $D$. 
The generator takes a vector of random numbers $\bm{z}$ as an input, and produces data with the same dimensions as the training data. 
On the other hand, the discriminator discriminates between samples from real data and data generated by the generator. 
The original version proposed by Goodfellow {\it et al.} \cite{Goodfellow2014} is called the vanilla GAN. 

In the training, $G$ and $D$ play the minimax game with the value function $V$ defined as follows: 
\begin{equation}
    \min_G \max_D V(D,G) \!=\! {\mathbb E}_{\bm{x} \sim p_{\rm data}(\bm{x})} \left[\log D(\bm{x})\right] \!+\! {\mathbb E}_{\bm{z} \sim p_z(\bm{z})} \left[ \log(1 \!-\! D(G(\bm{z})))\right]\!, 
\end{equation}
where $p_{\rm data}(\bm{x})$ and $p_z(\bm{z})$ are the distributions of the training data and $z$, respectively. 
The discriminator output $D(\bm{x})$ denotes the probability that $\bm{x}$ came from the real data distribution, 
and $G(\bm{z})$ represents a mapping from $\bm{z}$ to data space. 
This can be reformulated as the minimization of the Jensen--Shannon divergence 
between the real data and generated data distributions.

Following the proposal of the vanilla GAN, various derivations have been proposed. 
Major examples related to this study are described below. 

\subsection{Deep convolutional GAN}
The deep convolutional GAN (DCGAN) \cite{Radford2016} is a class of architectures of GANs 
based on convolutional neural networks (CNNs) mostly used for image generation tasks. 
In DCGAN, $G$ generates an image by repeating fractionally strided convolutions with a random number input $\bm{z}$.
The discriminator $D$ uses a CNN to infer whether the given image came from training data or data generated by $G$.

\subsection{Wasserstein GAN}
The Wasserstein GAN (WGAN) \cite{Arjovsky2017} is a variation of GANs 
that uses a metric different from the vanilla GAN.
WGAN defines the distance between distributions of training patterns and generated patterns based on the Wasserstein distance, and then minimizes it via training. 
This approach has the merit of stable learning with less mode collapse. 
In WGAN training, the minimax game is represented as follows: 
\begin{equation}
    \min_G \max_{D\in\mathcal{D}} V(D,G) 
    = {\mathbb E}_{\bm{x} \sim p_{\rm data}(\bm{x})} \left[ D(\bm{x})\right] 
	- {\mathbb E}_{\bm{z} \sim p_z(\bm{z})} \left[ D(G(\bm{z}))\right], 
\end{equation}
where $\mathcal{D}$ is a set of Lipschitz continuous functions. 
To satisfy the constraint that $D$ needs to be a Lipschitz function, 
$D$ is parameterized with weights lying in a compact space. 
Practically, the weights are clamped to a fixed box after each gradient update. 
For convenience, we call this method WGAN-Clipping in the rest of this paper.

The WGAN-gradient penalty (WGAN-GP) \cite{Gulrajani2017} is an improved version of the WGAN. 
The weight clipping performed in WGAN is an approximated approach as mentioned in the original paper. 
This approach often causes problems such as difficulty in adapting to a complicated distribution 
and inefficient learning with biased parameters. 
In WGAN-GP, to solve these problems, 
a gradient penalty is employed in the value function.
The training of WGAN-GP is expressed as follows: 
\begin{align}
	\min_G \max_D V(D,G) &= {\mathbb E}_{\bm{x} \sim p_{\rm data}(\bm{x})} \left[ D(\bm{x})\right] 
	- {\mathbb E}_{\bm{z} \sim p_z(\bm{z})} \left[ D(G(\bm{z}))\right] \nonumber \\	                              &\quad - \lambda \mathbb{E}_{\bm{\hat{x}} \sim {p}_{\hat{x}}(\bm{\hat{x}})} 
	\left[ ( {\| \nabla_{\bm{\hat{x}}} D(\bm{\hat{x}}) \|}_2 - 1)^2 \right], 
\end{align}
where $\bm{\hat{x}} = \epsilon\bm{x} + (1-\epsilon)G(\bm{z})$ 
and $\epsilon \sim U(0, 1)$. 
The WGAN-GP employs a new penalty term to make $D$ a Lipschitz function, 
thereby allowing more accurate and efficient learning compared with the original WGAN. 

\subsection{GANs with controlled output}\label{section:controllableGANs}
In ordinary GANs, it is difficult to predict what type of pattern 
will be generated from a certain input $z$ via the generator network. 
Many studies have therefore investigated the control of GANs' output. 
Mirza and Osindero \cite{Mirza2014} proposed the conditional GAN 
that can control the class of the generated image 
by adding class information encoded as a one-hot vector to the generator's input 
and a channel representing the class to the discriminator's input. 
Chen {\it et al.} \cite{Chen2016} proposed InfoGAN, 
where the generator's input is divided into information $c$ and noise $z$, 
and the discriminator is trained to discriminate not only between real and fake 
but also whether the generated data contain information of $c$. 
Odena {\it et al.} \cite{Odena2017} proposed AC-GAN with a strong constraint 
such that the class discrimination is also conducted in the discriminator
by adding class information to both the generator's and discriminator's inputs.  
Choi {\it et al.} \cite{StarGAN2018} used a domain label concatenated with the input image 
for multidomain image-to-image translation. 
Wang {\it et al.} \cite{wang2018high} proposed a GAN framework for synthesizing high-resolution images from semantic label maps. 
Shen {\it et al.} \cite{shen2018faceid} added the third network to the GAN framework to generate identity-preserving images. 
Liang {\it et al.} \cite{Liang_2018_ECCV} proposed contrast-GAN that modifies the semantic
meaning of an object by utilizing the object categories of both original and target domains. 
Bodla {\it et al.} \cite{bodla2018semi} achieved a semi-supervised approach by fusing an ordinary GAN and conditional GAN. 

There are also several GANs that share the parameters in the GAN model 
to have the same characteristics among multiple classes.
In \cite{Liu2016}, Liu and Tuzel proposed Coupled GAN 
where two GAN models learn different patterns 
while sharing some of the parameters, 
thereby generating a pair of patterns with similar tendencies. 
In addition, Mao {\it et al.} \cite{Mao2017} proposed AlignGAN, 
which can control the domain and class of the generated data with a consistent pattern. 
	
	\section{GlyphGAN: Style-Consistent Font Generation}
	Figure \ref{05_training_ours} shows an overview of GlyphGAN. 
The major differences from the ordinary GANs are as follows.
\begin{itemize}
	\item The input vector $\bm{z}$ of the generator $G$ consists of a style vector $\bm{z}^{\rm s}$ and a character class vector $\bm{z}^{\rm c}$. 
	\item During training, the character class vector $\bm{z}^{\rm c}$ is associated with the character class of the training pattern. 
\end{itemize}
\begin{figure*}[t]
	 \begin{minipage}{\hsize}
	    \centering
	    \includegraphics[width=\hsize]{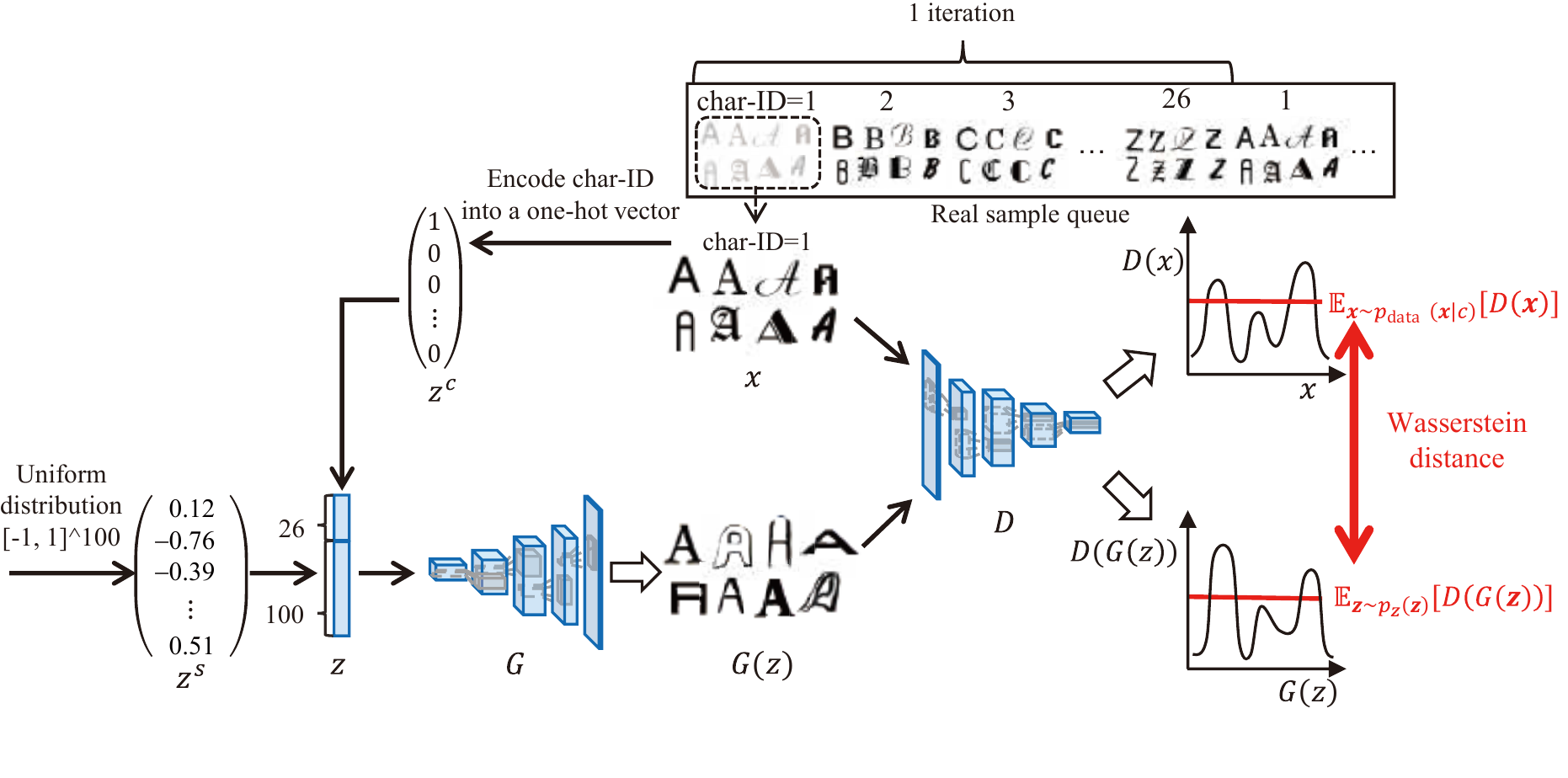}
	 \end{minipage}
	\caption{Overview of GlyphGAN. 
		The generator $G$, which generates synthetic fonts similar to manually-designed ones, 
		and the discriminator $D$, which discriminates between generated and real fonts, 
		are adversarially trained. 
		In GlyphGAN, the input vector $\bm{z}$ for $G$ consists of the style vector $\bm{z}^{\rm s}$ 
		and character class vector $\bm{z}^{\rm c}$. 
		The style vector $\bm{z}^{\rm s}$ is a uniform random vector, and the character class vector $\bm{z}^{\rm c}$ is a one-hot vector associated with the character class of training patterns. 
		The discriminator $D$ calculates the distance between 
		distributions of generated data and training data based on the Wasserstein distance. 
		Through learning, the parameters of $G$ are optimized to minimize this distance.}
	\label{05_training_ours}
\end{figure*}

\subsection{Input vector}
Let $\bm{z}$ be the input of the generator $G$. 
In GlyphGAN, $\bm{z}$ consists of a style vector $\bm{z}^{\rm s}$ 
and a character class vector $\bm{z}^{\rm c}$. 
By independently preparing input vectors for the style and character class, 
various character images can be generated with the style fixed, and vice versa. 

Let the style vector $\bm{z}^{\rm s}$ be a 100-dimensional random number sampled from a uniform distribution. 
This is the same setting as the ordinary GANs. 

Let the character class vector $\bm{z}^{\rm c}$ be a one-hot vector corresponding to the character class. 
Taking the alphabet for example as shown in Fig. \ref{05_training_ours}, 
the character IDs such as 1, 2, 3, $\ldots$ associated with the character classes ``A,'' ``B,'' ``C,'' $\ldots$
are encoded to the one-hot format. 
The number of dimensions of $\bm{z}^{\rm c}$ is the total number of characters used for learning. 
For example, it is 26 for upper-case Latin alphabets.

\subsection{Network architecture}
GlyphGAN basically employs DCGAN's network architecture \cite{Radford2016}. 
The generator $G$ takes a random vector as an input and then outputs an image with the same size as the training images. 
Each layer of $G$ is a fractionally strided convolution. 
ReLU activation is used except for the output layer that uses Sigmoid.
The discriminator $D$ takes an image and outputs a scalar value. 
Each layer is a strided convolution, instead of using a pooling layer with an ordinary convolution layer. 
LeakyReLU was applied to each layer of $D$. 
Different from the original DCGAN, GlyphGAN does not employ batch normalization, 
following the recommendation in \cite{Gulrajani2017}.

\subsection{Training algorithm}
Algorithm \ref{al:LearningAlgorithm} shows the training algorithm of GlyphGAN.
\begin{algorithm}[ht]
\caption{Training algorithm of GlyphGAN}
\label{al:LearningAlgorithm}
\SetAlgoLined
\SetKwInOut{Input}{Require}
\Input{Coefficient $\lambda$, number of discriminator iterations per generator iteration $N_{\rm disc}$, batch size $M$, Adam hyperparameters $\alpha, \beta_1, \beta_2$}
Initialize discriminator's parameters $w$ and generator's parameters $\theta$
    \For{{\rm number of training epochs}}{
        \For{$c = 1, \ldots, 26$}{
            Set one-hot vector $\bm{z}^{\rm c}$ corresponding to character class $c$\;
            \For{$t = 1, \ldots, N_{\rm disc}$}{
                \For{$i = 1, \ldots, M$}{
                    Sample real data $\bm{x} \sim p_{\rm data}(\bm{x}|c)$, style vector $\bm{z}^{\rm s} \sim p_z(\bm{z}^{\rm s})$, a random number $\epsilon \sim U(0, 1)$\;
                    $\bm{z} \gets ({\bm{z}^{\rm s}}^{\rm T}, {\bm{z}^{\rm c}}^{\rm T})^{\rm T}$\;
                    $\bm{\hat{x}} \gets \epsilon\bm{x} + (1-\epsilon)G_{\theta}(\bm{z})$\;
                    $L^{(i)} \gets D_w(G_{\theta}(\bm{z}))$  - $D_w(\bm{x}) + \lambda ({\| \nabla_{\bm{\hat{x}}} D_w(\bm{\hat{x}}) \|}_2 - 1)^2$\;
                }
                $w \gets \mathrm{Adam}(\nabla_w \frac{1}{m}\sum^{m}_{i=1}L^{(i)}, w, \alpha, \beta_1, \beta_2)$
            }
            Sample a batch of style vectors  $\{\bm{z}^{\rm s}_i\}^M_{i=1} \sim p_z(\bm{z}^{\rm s})$\;
            $\bm{z}_i \gets ({\bm{z}^{\rm s}_i}^{\rm T}, {\bm{z}^{\rm c}}^{\rm T})^{\rm T}$\;
            $\theta \gets \mathrm{Adam}(\nabla_\theta \frac{1}{m}\sum^{m}_{i=1}-D_w(G_{\theta}(\bm{z}_i)), \theta, \alpha, \beta_1, \beta_2)$
        }
    }
\end{algorithm}
The training algorithm of GlyphGAN basically follows that of WGAN-GP. 
Different from WGAN-GP's algorithm, a one-hot vector $\bm{z}^{\rm c}$ representing the character class is embedded into the latent vector, and is associated with the character ID of the training data.
Given a set of font images $\bm{x}$ with a character class $c = 1, \ldots, 26$ for each image, the networks are trained. 
First, with a fixed $\bm{z}^{\rm c}$, only the corresponding characters are used. 
For example, Fig. \ref{05_training_ours} illustrates the stage of learning ``A.'' 
The style vector $\bm{z}^{\rm s}$ is sampled from a uniform distribution 
and then concatenated with $\bm{z}^{\rm c}$ to make the generator's input $\bm{z}$. 
The networks $G$ and $D$ are trained using $\bm{z}$ and the images of the corresponding character class. 
In this stage, we use only a batch of images randomly selected from all of the training images. 

After learning with respect to one character class, 
we move on to the learning of the next character class. 
In the example of Fig. \ref{05_training_ours}, ``B'' becomes the next target. 
After that, we proceed to the learning of ``C,'' ``D,'' ``E,'' ..., ``Z,'' continuously, and then return to ``A.'' 
By learning repeatedly for each character class in this way, 
we prevent the network from overfitting to a specific character class. 
A series of learning for all of the character classes is counted as an epoch.

In the training of GlyphGAN, the WGAN-GP-based value function \cite{Gulrajani2017} is used. 
Since the data sampling procedure and the input vector are different from the original WGAN-GP, 
the minimax game is reformulated as follows: 
\begin{align}
	\min_G \max_D V(D,G) &= {\mathbb E}_{\bm{x} \sim p_{\rm data}(\bm{x}|c)} \left[ D(\bm{x})\right] 
	- {\mathbb E}_{\bm{z}^{\rm s} \sim p_z(\bm{z}^{\rm s})} \left[ D(G(\bm{z}))\right] \nonumber \\
	&\quad - \lambda {\mathbb E}_{\bm{\hat{x}} \sim {p}_{\hat{x}}(\bm{\hat{x}})} 
	\left[ ( {\| \nabla_{\bm{\hat{x}}} D(\bm{\hat{x}}) \|}_2 - 1)^2 \right], 
\end{align}
where $\bm{z} = ({\bm{z}^{\rm s}}^{\rm T}, {\bm{z}^{\rm c}}^{\rm T})^{\rm T}$,  
$c$ is the character class corresponding to $\bm{z}^{\rm c}$, 
$\bm{\hat{x}} = \epsilon\bm{x} + (1-\epsilon)G(\bm{z})$, 
and $\epsilon \sim U(0, 1)$. 
	
	\section{Font Generation Experiment}
	To evaluate the capability of the proposed method, 
we conducted a font generation experiment. 
We evaluated the generated fonts from the following viewpoints: 
\begin{itemize}
	\item Legibility: We verify that the generated font has legibility via a character recognition experiment using a pretrained CNN.
	\item Diversity: We validate whether the generated font set has diversity different from the training data. 
	\item Style consistency: We qualitatively verify that the generated font has style consistency via visual observation, and then quantitatively evaluate the effect of a training data shortage on style consistency.
\end{itemize}

\subsection{Dataset}
Figure \ref{05_real_30fonts} shows 30 examples randomly selected from the dataset. 
For the dataset, we prepared 26 uppercase alphabet letters from 6,561 different fonts.
Each image was a grayscale image with a size of $64 \times 64$. 
Although the sizes of the prepared fonts differed slightly from each other 
even if we set the same number of points, 
we used them without normalization, regarding it as one of the font features. 
\begin{figure}[!t]
	\begin{minipage}{\hsize}
		\centering
		\includegraphics[width=0.8\hsize]{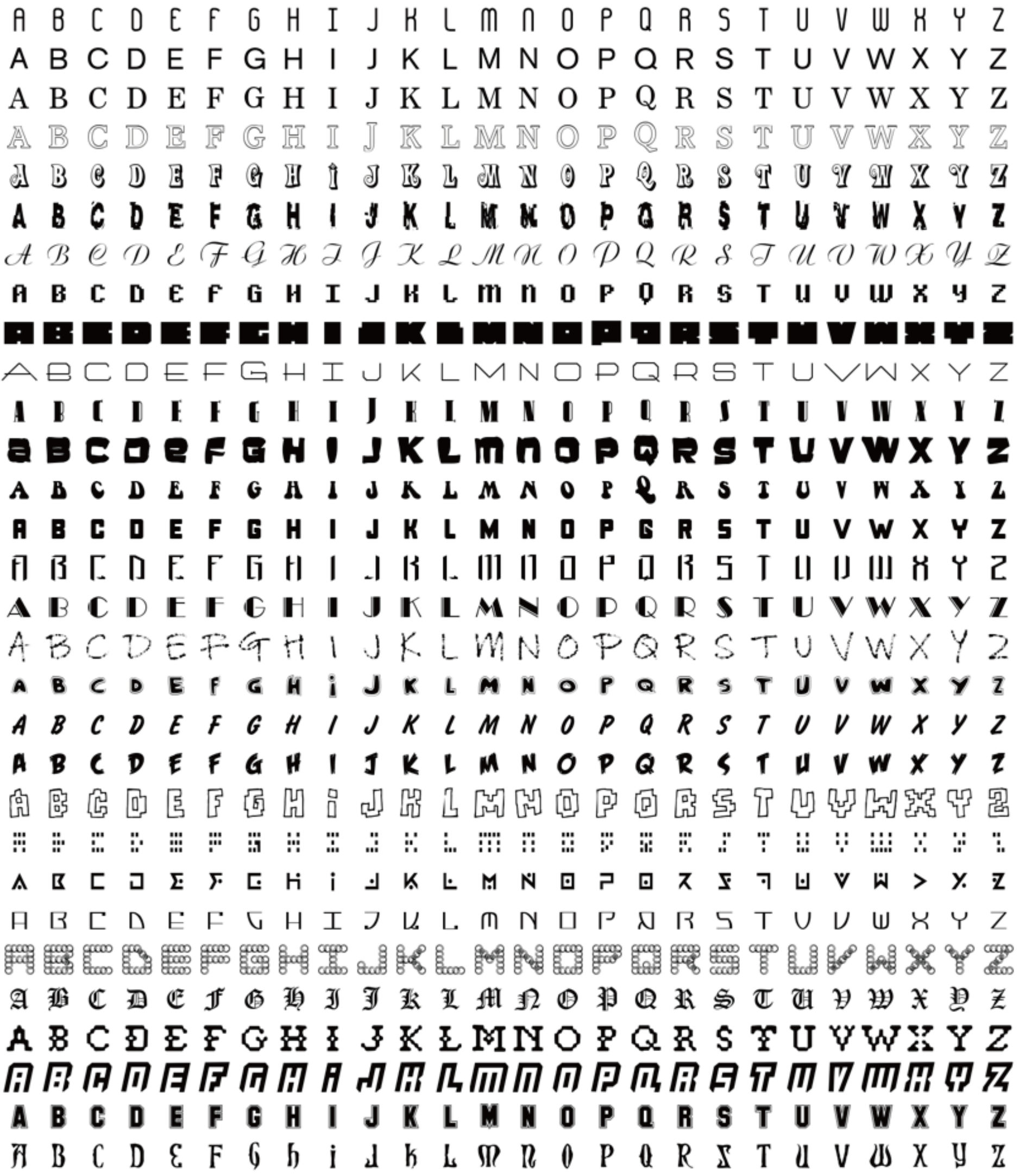}
	\end{minipage}
	\caption{Examples of training patterns used in the experiment.}
	\label{05_real_30fonts}
\end{figure}

\subsection{Details of the Network Structure and Parameter Settings}
Figure \ref{05_generator_discriminator} shows the structures of $G$ and $D$ used in the experiment. 
Basically, these networks are the same as DCGAN \cite{Radford2016} with the image size adjusted. 
For the activation functions of $G$, 
we used ReLU except for the final layer that employed Sigmoid. 
LeakyReLU was applied to each layer of $D$. 
The algorithm of the gradient descent method and its hyperparameters were determined
according to \cite{Gulrajani2017}.
For weight updating, we used Adam \cite{Kingma2015} with the parameters of $\alpha = 0.0002$, $\beta_1 = 0.5$, and $\beta_2 = 0.99$.
Batch normalization was not applied. 
The number of discriminator iterations per generator iteration was set as $N_\mathrm{disc}=5$
The number of learning iterations was 2,500.
The batch size was set as 1,024.
\begin{figure}[!t]
    \centering
    \includegraphics[width=\hsize]{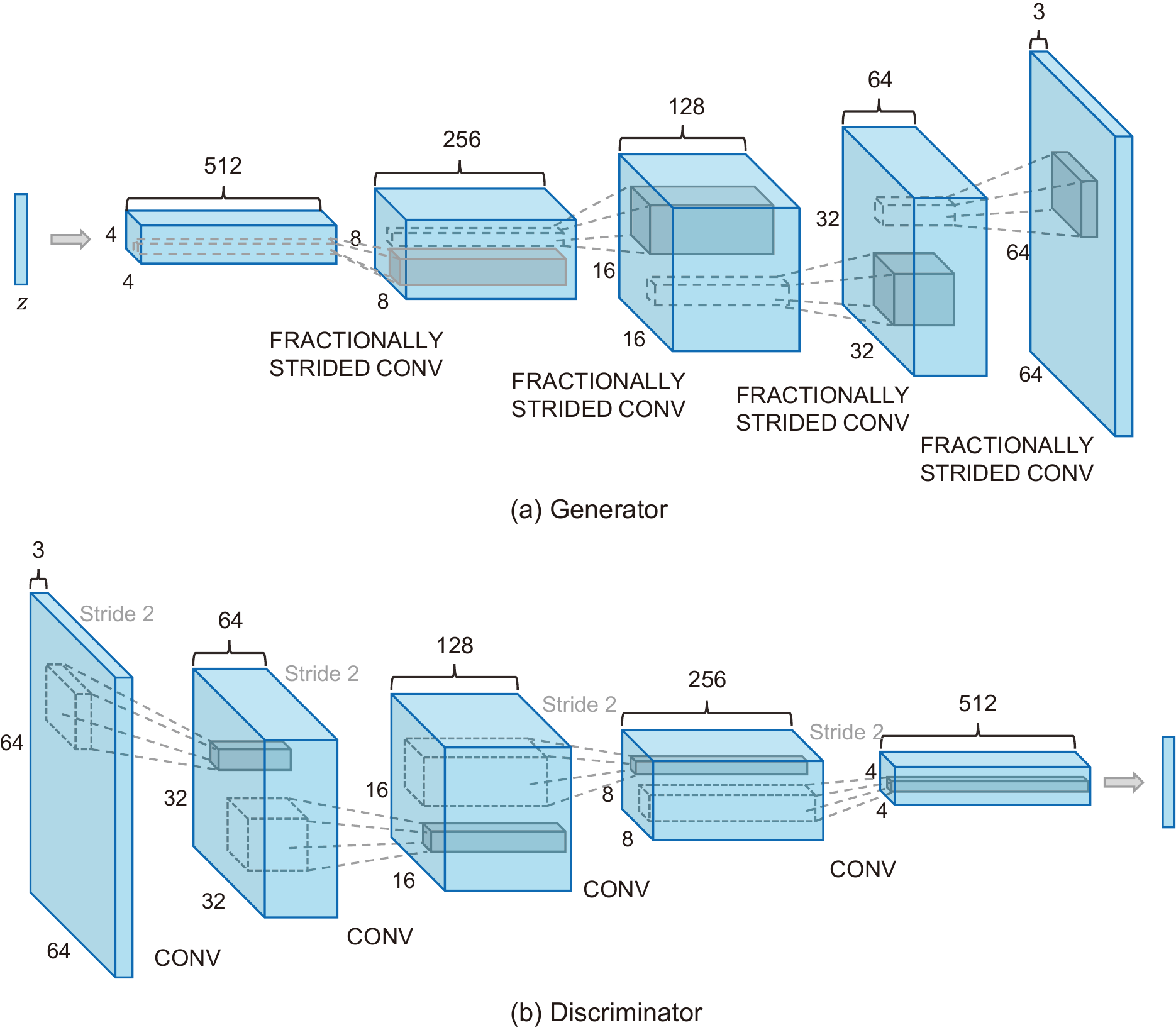}
	\caption{Structures of the generator and the discriminator.}
	\label{05_generator_discriminator}
\end{figure}

\subsection{Generation Results}
Figure \ref{05_result_wgan-gp} shows the generation results; 
examples of generated fonts with randomly selected style vectors $\bm{z}^{\rm s}$. 
The results generated by changing character class vector $\bm{z}^{\rm c}$ with a fixed style vector $\bm{z}^{\rm s}$ are aligned horizontally. 
The results with different style vectors and a fixed character class vector are aligned vertically. 
In each line, the generated letters have a similar font style consistently over all characters 
even though they are independently generated with the same $\bm{z}^{\rm s}$. 
In addition, by changing $\bm{z}^{\rm s}$, GlyphGAN generated fonts with various types of styles 
such as serif, sans-serif, thickness, roundness, and size, even including a font made of outlines. 
\begin{figure}[!t]
	 \begin{minipage}{\hsize}
	    \centering
	    \includegraphics[width=0.8\hsize]{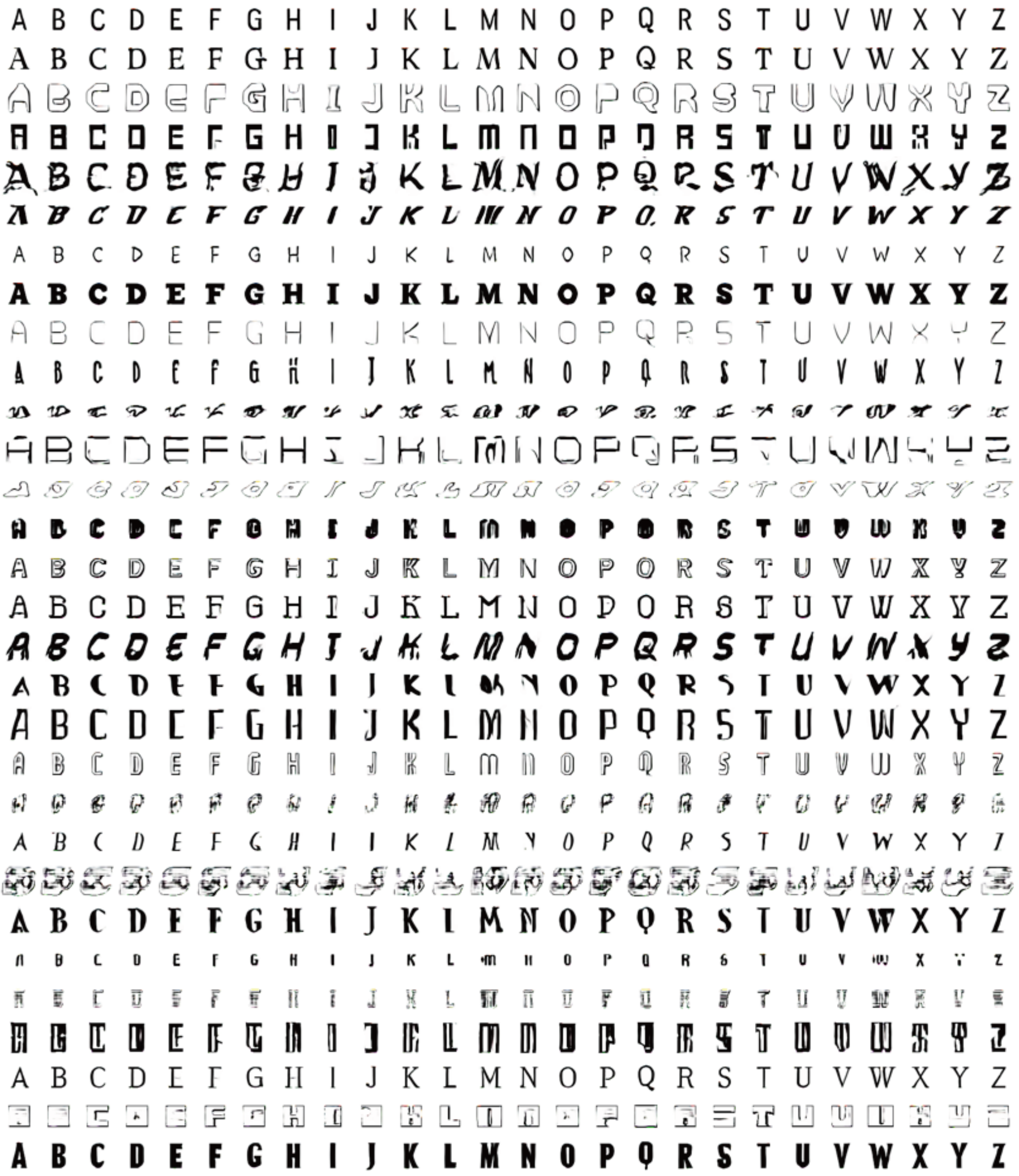}
	 \end{minipage}
	\caption{Thirty randomly selected examples from the generated font sets.
		It should be emphasized that letters have similar font style consistently over all alphabets 
		(``A'' to ``Z''), although they are generated independently but with the same $\bm{z}^{\rm s}$. }
	\label{05_result_wgan-gp}
\end{figure}

Figure \ref{05_result_walk_A} shows 
the letter ``A'' generated with a continuously changing $\bm{z}^{\rm s}$. 
In this result, 128 points were randomly selected from the $\bm{z}^{\rm s}$ space. 
The vector $\bm{z}^{\rm s}$ was then moved along eight points that linearly interpolated every two points out of the 128 points.  
The style of generated font smoothly changed according to the move of style vector $\bm{z}^{\rm s}$, 
demonstrating the possibility of fine control of generated styles. 
This result shows the capability of generating an intermediate font between the existing two fonts. 
\begin{figure}[!t]
	 \begin{minipage}{\hsize}
	    \centering
	    \includegraphics[width=0.9\hsize]{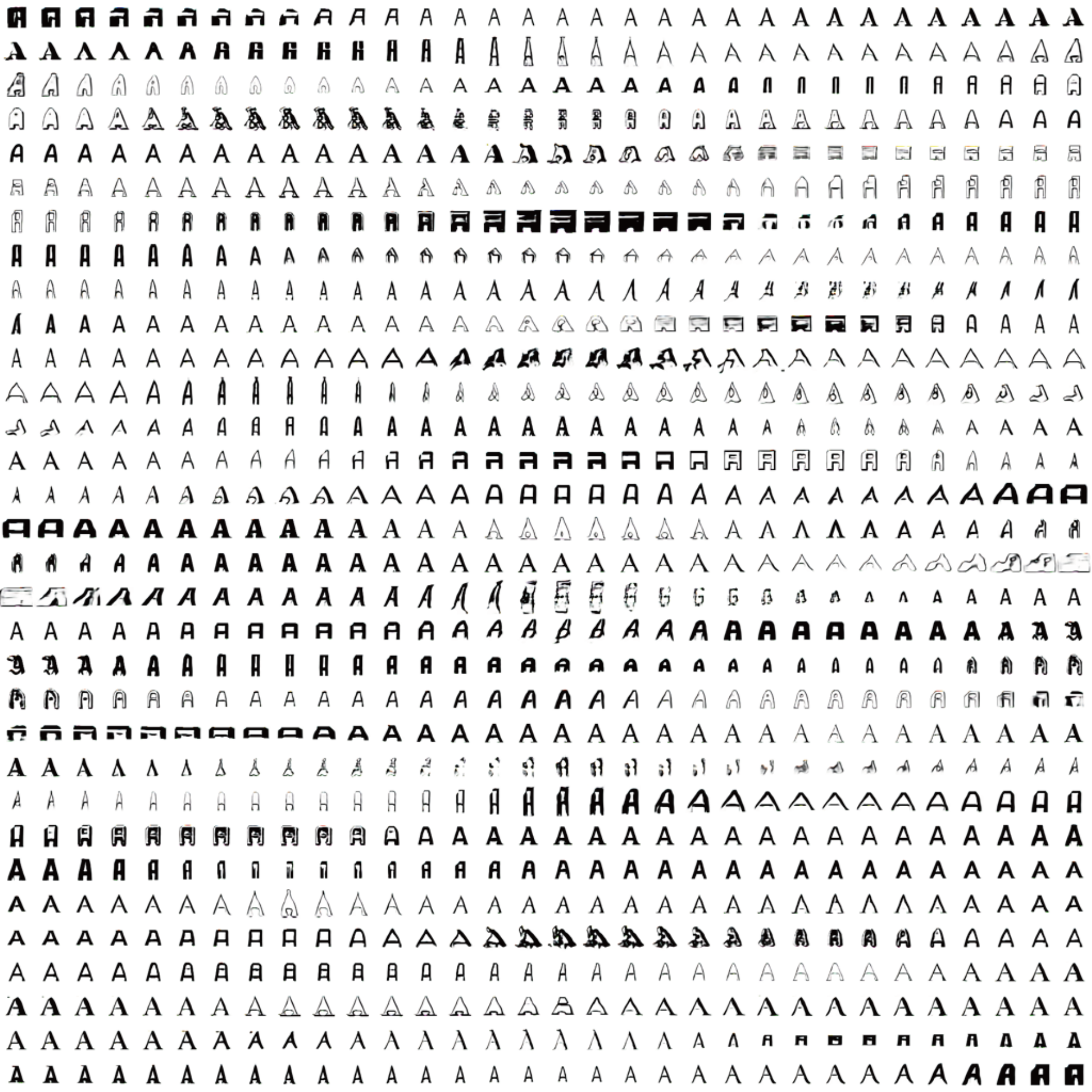}
	 \end{minipage}
	\caption{Generated results with a continuously changing $z^s$ for the letter ``A''}
	\label{05_result_walk_A}
\end{figure}

\subsection{Legibility Evaluation}
\label{subsection:legibility}
To evaluate the legibility of the generated fonts, 
a recognition experiment was performed using a multi-font character recognition CNN. 
Legibility is indispensable in font generation and is also one of the important indicators in this research. 

Figure \ref{05_classifier} shows the structure of the multi-font character recognition CNN used in this experiment. 
This CNN consisted of four convolutional layers and a fully-connected layer. 
ReLU was employed as an activation function after each layer. 
In addition, batch normalization and dropout were applied after the fully-connected layer. 
\begin{figure}[!t]
	 \begin{minipage}{\hsize}
	    \centering
	    \includegraphics[width=\hsize]{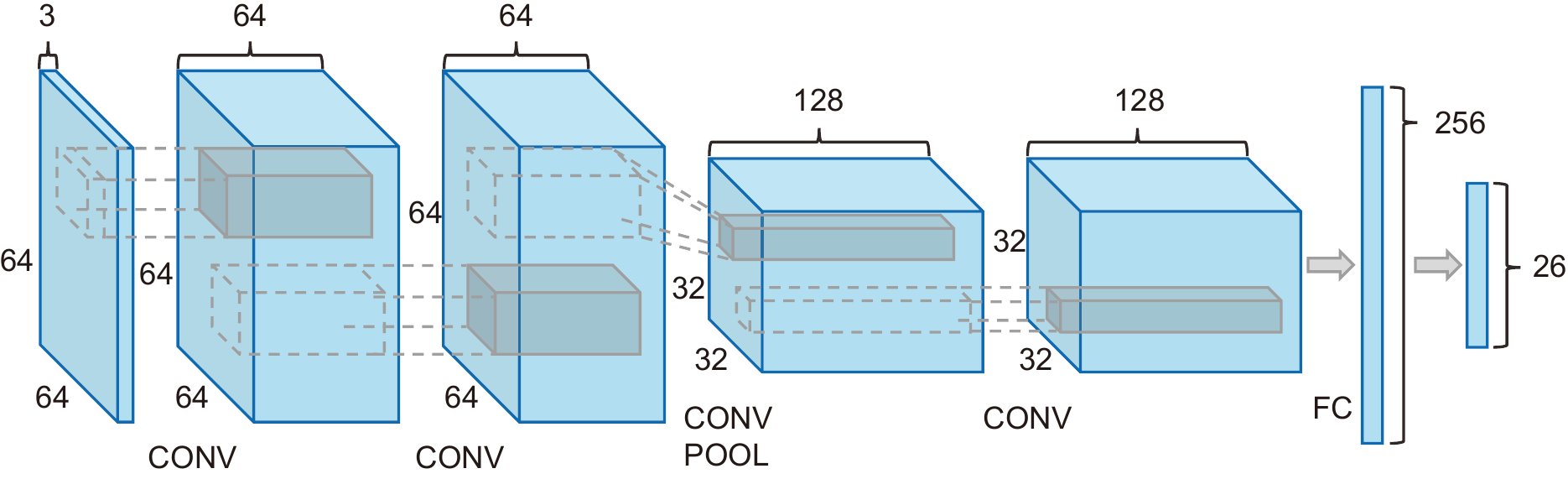}
	 \end{minipage}
	\caption{Structure of the multi-font character recognition CNN used in the legibility evaluation.}
	\label{05_classifier}
\end{figure}

We confirmed the basic ability of this CNN for character recognition using existing fonts. 
By dividing 6,561 fonts of 26 alphabet capital letters 
into the training and testing sets with the ratio of 9:1, 
the CNN was trained to classify 26 letters. 
As a result, the training accuracy and testing accuracy were $98.27$ \% and $93.26$ \%, respectively. 

In the evaluation, we generated 10,000 fonts with 26 uppercase letters using GlyphGAN. 
For comparison, we replaced the value function for the learning of GlyphGAN 
with those of DCGAN \cite{Radford2016} and WGAN-Clipping \cite{Arjovsky2017}, 
and generated 10,000 fonts for each comparative method. 

Figure \ref{05_result_change_framework} shows examples of the comparative methods. 
Table \ref{05_result_recognition} shows the results of the character recognition.
Compared with DCGAN and WGAN-Clipping, 
the recognition accuracy for the generated fonts by GlyphGAN was higher, 
thereby showing the effectiveness of learning GlyphGAN for improving the legibility of the generated fonts. 
\begin{figure}[!t]
	\centering
	\includegraphics[width=0.8\hsize]{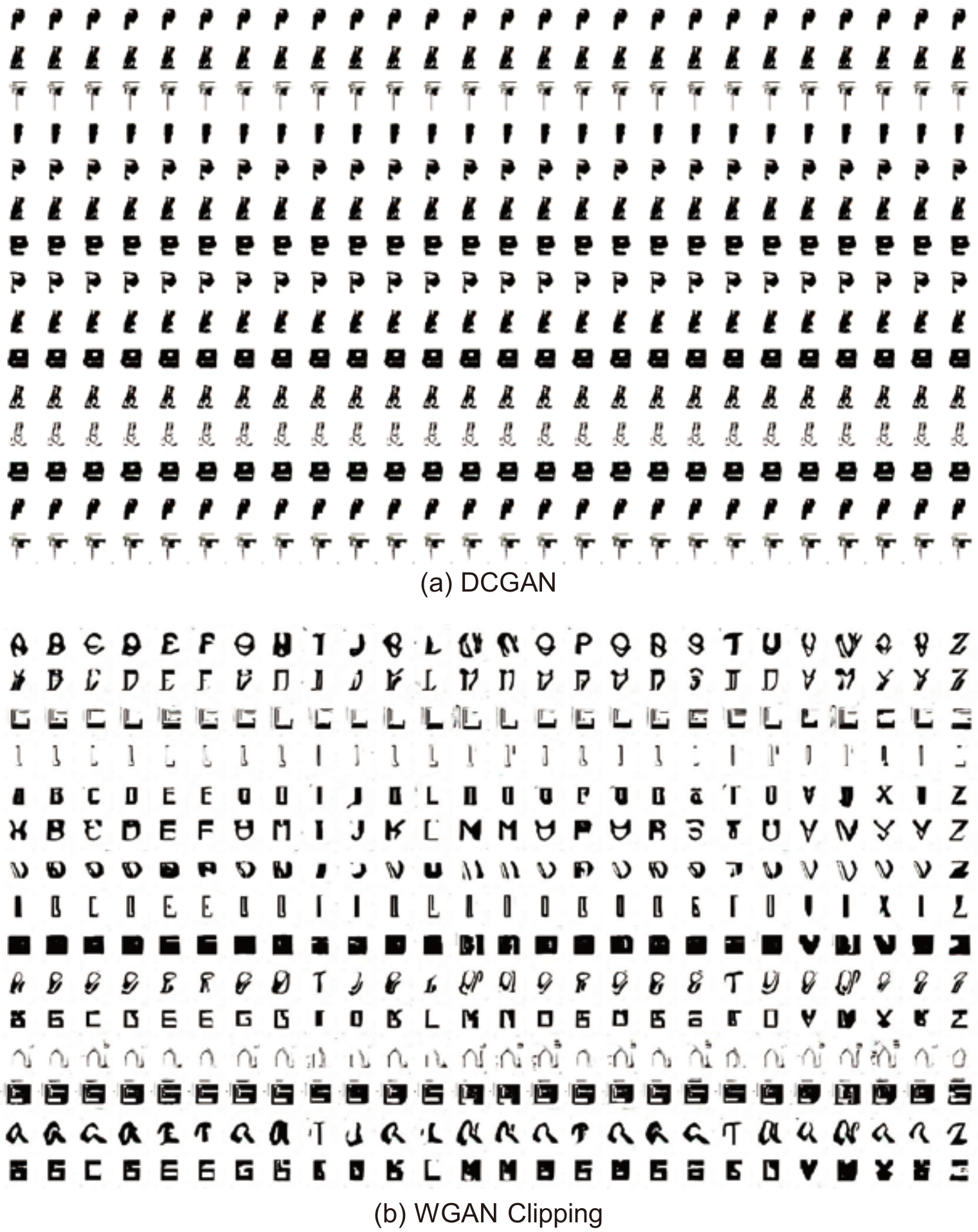}
	\caption{Generated fonts using different learning algorithms}
	\label{05_result_change_framework}
\end{figure}
\begin{table}[t]
	\centering
	\caption{Results of character recognition for legibility evaluation.}
	\label{05_result_recognition}
	\begin{tabular}{ll}
		\hline
		Method			& Accuracy {[}\%{]}	\\ \hline
		DCGAN			& 3.90				\\
		WGAN-Clipping	& 32.92				\\
		GlyphGAN (ours)	& 83.90				\\ \hline
	\end{tabular}
\end{table}

\subsection{Diversity Evaluation}
In this evaluation, we validated whether the generated font set had diversity, 
that is, the generated fonts were different from the training patterns. 
The generated fonts are sometimes similar to the font used as a training pattern. 
In a sense, it is reasonable to have similar fonts because the goal of GANs is to reproduce the training patterns. 
On the other hand, there can be unknown patterns that are not seen in the training patterns
if GlyphGAN can estimate the mapping from the distribution of $\bm{z}^{\rm s}$ onto the manifold that is constructed by the training patterns. 

Figure \ref{05_tendency_method} shows an outline of the analysis method used in this evaluation. 
We analyzed the tendency of the generated patterns by measuring the distance between the generated patterns and training patterns.
Using 10,000 $\times$ 26 generated patterns which are the same as in the legibility evaluation, 
we calculated the minimum value among pseudo-Hamming distances \cite{Uchida2015} between each generated pattern and training patterns in the corresponding character class. 
We then define the distance between the generated patterns and the nearest training patterns for each style
as the average of the minimum values over all of the character classes. 
We also defined the most similar font as 
an existing font to which the minimum value was most frequently assigned. 
\begin{figure}[!t]
    \centering
    \includegraphics[width=\hsize]{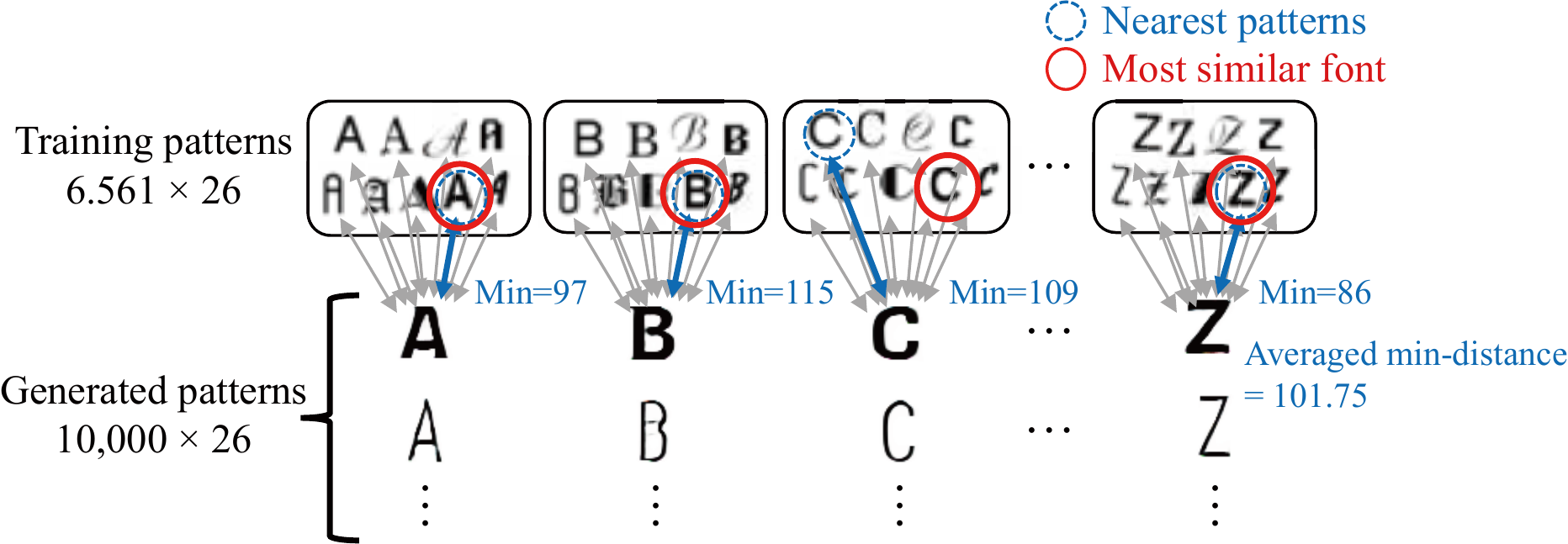}
	\caption{Outline of the measurement method of the distance between generated and training patterns. 
		The distance between each generated and training pattern is calculated based on the pseudo-Hamming distance. 
		The training pattern nearest to the generated pattern is surrounded by a blue dotted circle. 
		The distance between generated patterns and the nearest training patterns is defined
		as an average of the minimum distances over all the character classes. 
		The red circles indicate the most similar font, which is an existing font to which the minimum distance is most frequently assigned. }
	\label{05_tendency_method}
\end{figure}

Figure \ref{05_result_histogram} shows a histogram of the distances between generated patterns and the nearest training patterns. 
The minimum and maximum distances were 40.30 and 2942.86 (0.98 \% and 71.84 \% of the total pixel number), respectively. 
In addition, the generated patterns with a distance of less than 500 accounted for 87.51 \%. 
\begin{figure}[!t]
    \centering
    \includegraphics[width=\hsize]{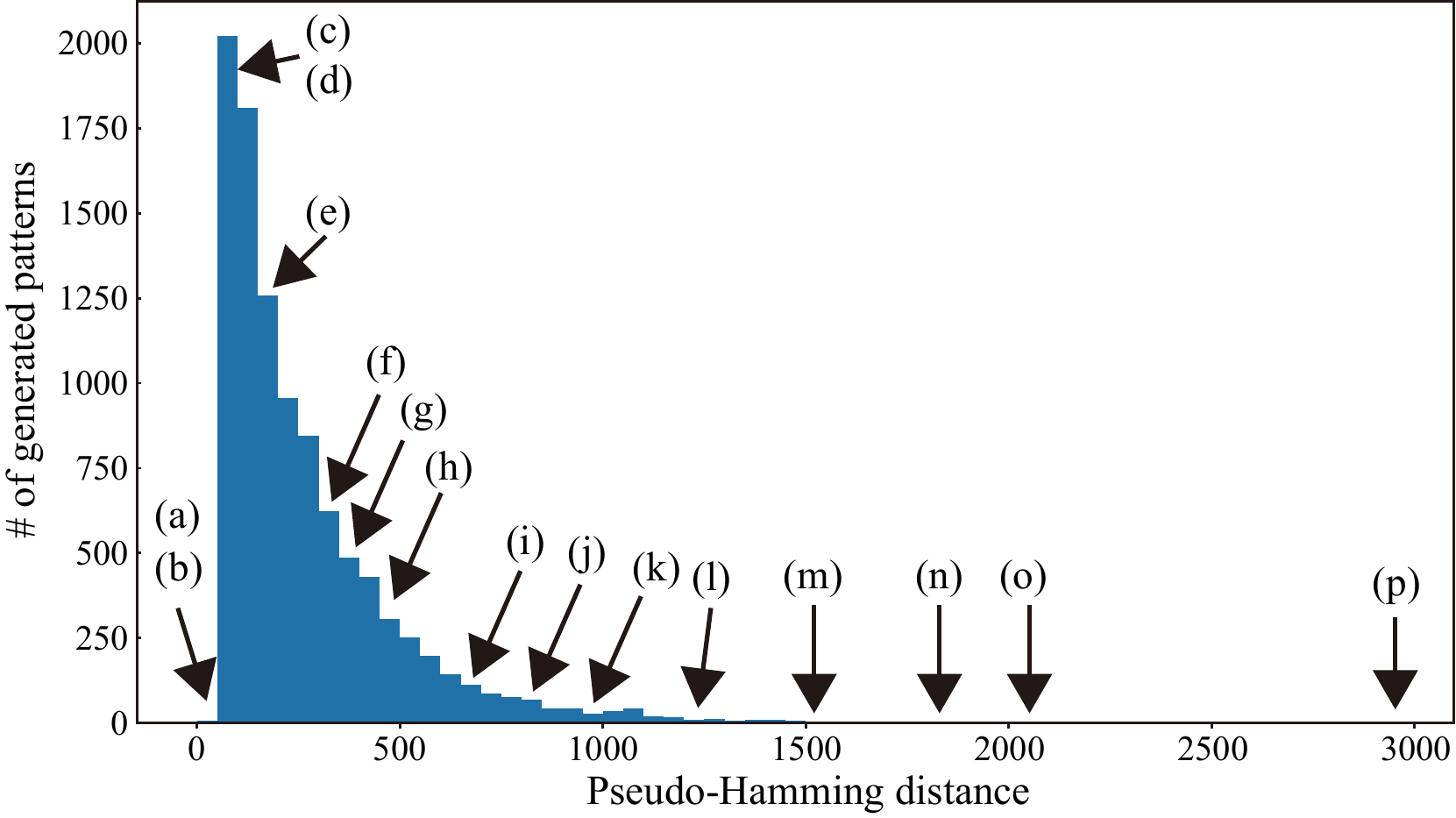}
    \caption{Histogram of distances between generated patterns and the nearest training patterns calculated based on the pseudo-Hamming distance. 
  	The points indicated by the arrows correspond to subfigures in Figure \ref{05_result_tendency}.}
    \label{05_result_histogram}
\end{figure}

Figure \ref{05_result_tendency} shows examples of the generated patterns and the most similar font in the training patterns. 
In the examples with small distances such as in Fig. \ref{05_result_tendency}(a), 
fonts that look similar to the training pattern are observed. 
On the other hand, in the examples with large distances, 
the generated patterns are greatly different from the training patterns. 
These examples can be regarded as styles not seen in the training patterns. 
Although such patterns are relatively few 
(the ratio of samples with a distance greater than 500 is about 10 \% of the total), 
the font set generated by the proposed method has diversity different from the training patterns.
\begin{figure}[!t]
    \centering
	\includegraphics[width=0.8\hsize]{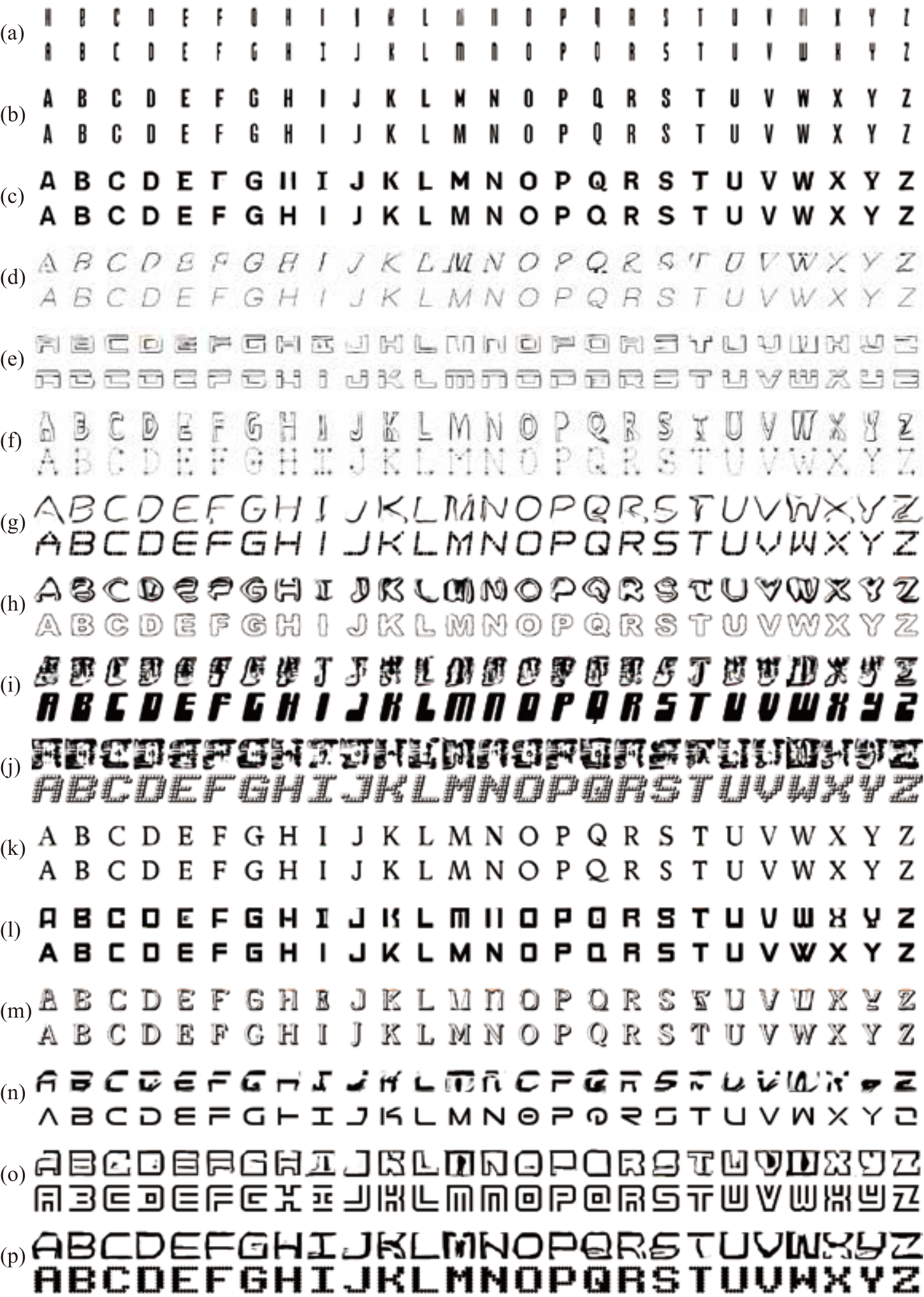}
	\caption{Comparison of generated patterns and the most similar font in the training patterns. 
		In each subfigure, the top shows generated ones and the bottom shows training ones.}
	\label{05_result_tendency}
\end{figure}

\subsection{Effect of Training Data Shortage on Style Consistency}
\label{section:shortage}
We explore the effect of a training data shortage on style consistency. 
From original 6,561 fonts, we gradually decreased the number of fonts in the training dataset 
as 1,000, 100, 10 by randomly selecting fonts from the original font set. 
After training GlyphGAN with each selected font set, 
we quantitatively evaluated the style consistency of the generated font images. 
For quantitative evaluation, we defined the metric of style consistency by 
\begin{equation}
    C_\mathrm{s} = \frac{1}{NC}\sum^N_{n=1}\frac{1}{\bar{d}_{n}}\sum^C_{c=1}(d_{n,c} - \bar{d}_{n})^2, 
\end{equation}
where $N$ is the number of generated images (we used $N=10,000$ in this experiment), $C$ is the number of character classes, 
i.e., $C = 26$, $d_{n,c}$ is the distance between the generated font and the nearest real font used in Section 5.5, and $\bar{d}_{n}$ is the average of $d_{n,c}$ over $c$. 
The metric $C_\mathrm{s}$ is the averaged coefficient of variation of $d_{n,c}$, 
and represents an intra-style variation of the generated font images. 
The lower $C_\mathrm{s}$ is, the higher style consistency is.

As an example of the generated fonts with a limited training font set, 
generated results with the training dataset having only 10 fonts are presented 
in Figure \ref{05_result_only10}.
Compared with the result in Fig. \ref{05_result_wgan-gp}, style consistency is not maintained. 
\begin{figure}[!t]
		\begin{minipage}{\hsize}
			\centering
			\includegraphics[width=\hsize]{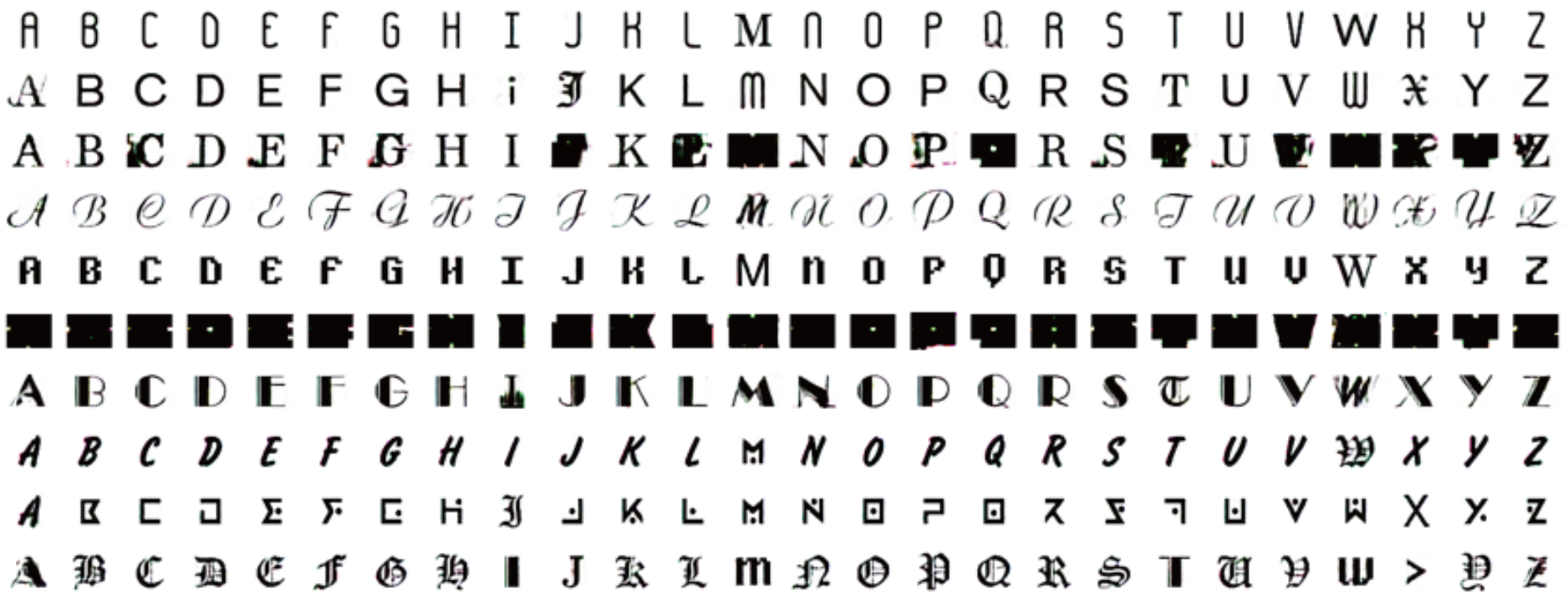}
		\end{minipage}
	\caption{Generated results with the training dataset having only 10 fonts.}
	\label{05_result_only10}
\end{figure}
Table \ref{Consistency} shows the relationships between the number of training font and the metric for style consistency $C_\mathrm{s}$. 
\begin{table}[t]
	\centering
	\caption{Effect of the number of training fonts on style consistency.}
	\label{Consistency}
	\begin{tabular}{ll}
		\hline
		Number of training fonts    & $C_\mathrm{s}$	\\ \hline
		10			                & 1.12				\\
		100     	                & 0.51				\\
		1000     	                & 0.47				\\
		6561    	                & 0.46				\\ \hline
	\end{tabular}
\end{table}
The metric $C_\mathrm{s}$ increases according to the decrease in the number of training fonts.
These results suggest that a sufficient number of styles 
is required for the training data to guarantee style consistency. 

\subsection{Quantitative Comparison with the Existing Method}
We conducted a quantitative comparison with deep-fonts \cite{Bernhardsson2016blog, Bernhardsson2016git}.
Deep-fonts is a neural network-based generative model for font images. 
As with GlyphGAN, deep-fonts takes the concatenated vector of a random vector representing a style and a one-hot vector representing a character class as input and outputs a font image.
The main differences of deep-fonts from GlyphGAN are as follows:
\begin{itemize}
    \item Network structure: A multilayer perceptron-based network is employed instead of a CNN.
    \item Loss function: Generated font images are evaluated by $L_1$ loss between the generated and real font images instead of using a discriminator.
\end{itemize}

Figure \ref{deep-fonts} shows the example font images generated by deep-fonts. 
\begin{figure}[t]
		\begin{minipage}{\hsize}
			\centering
			\includegraphics[width=\hsize]{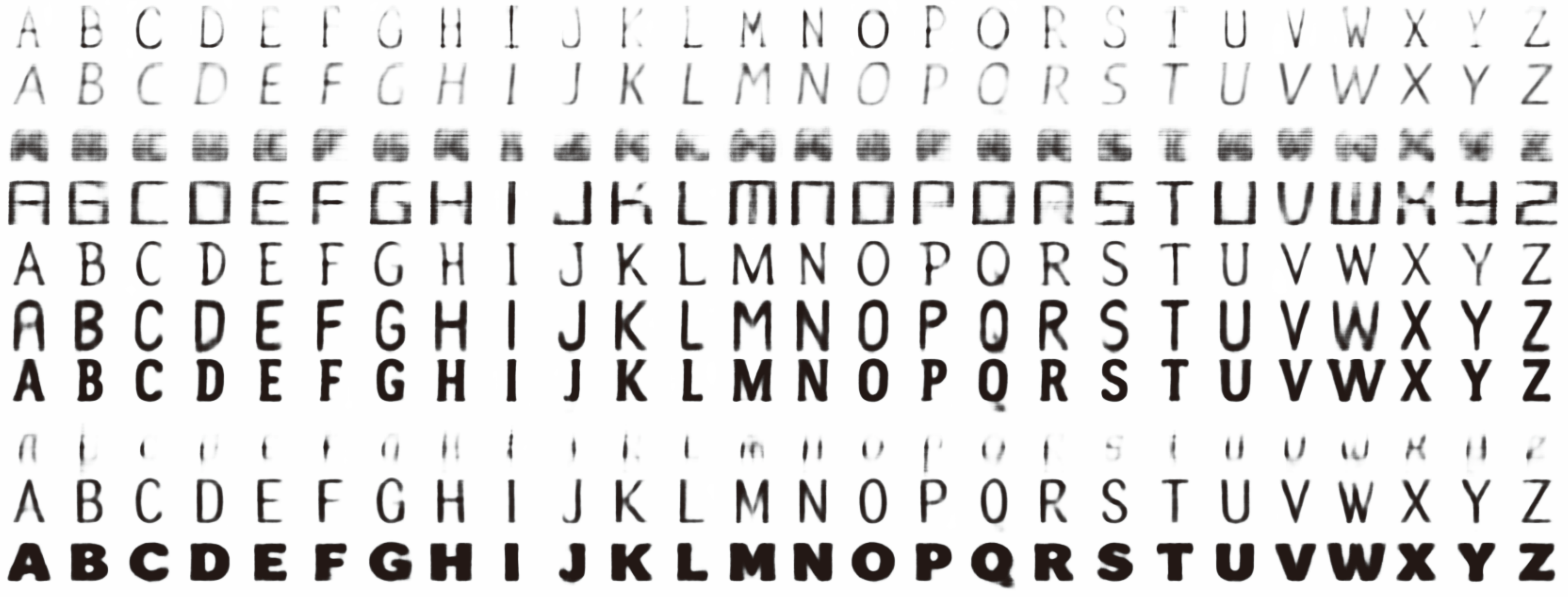}
		\end{minipage}
	\caption{Generated results using deep-fonts.}
	\label{deep-fonts}
\end{figure}
It seems that the generated fonts are legible, and have style consistency and diversity to some extent. However, there are some collapsed fonts such as third and eighth rows. 

We compared the qualities of the generated fonts by GlyphGAN and deep-fonts 
in terms of legibility, style consistency, and diversity. 
As the metrics for legibility and style consistency, we employed recognition accuracy and $C_\mathrm{s}$ used in Sections 5.4 and 5.6, respectively. 
Furthermore, We defined the metric of diversity as follows:
\begin{equation}
    C_\mathrm{d} = \frac{1}{2Q_2}(Q_3 - Q_1),
\end{equation}
where $Q_1$ and $Q_3$ are the first and third quartiles of $\{\bar{d}_n\}^N_{n=1}$, and $Q_2$ is a median.
This metric intuitively represents the coefficient of deviation of Figure \ref{05_result_histogram}. 
Since there are some outliers due to collapsed generated images, 
we used quartile deviation instead of standard deviation. 

The results of the quantitative comparison are shown in Table \ref{qualitative_comparison}. 
\begin{table}[t]
	\centering
	\caption{Quantitative comparison with Deep-fonts.}
	\label{qualitative_comparison}
	\scalebox{0.9}{ 
    	\begin{tabular}{llll}
    		\hline
    		                & Recognition accuracy {[}\%{]} & $C_\mathrm{s}$ & $C_\mathrm{d}$ \\
    		Method		& (Legibility)      & (Style consistency)    & (Diversity) \\ \hline
    		Deep-fonts	& 72.51		        & 0.47             & 0.33    \\
    		GlyphGAN	& \textbf{83.90}	& \textbf{0.46}    & \textbf{0.61}    \\ \hline
    	\end{tabular}
	}
\end{table}
In particular, GlyphGAN showed better legibility than deep-fonts. 
This is because deep-fonts occasionally generates collapsed and illegible font images. 
Introducing the GAN framework allowed the generator to estimate a smooth font manifold, 
thereby improving the style consistency and diversity of the generated fonts.

	
	\section{Discussion}
	Difference from existing GANs:
As stated in the Related Work section, 
various GAN derivations have been proposed. 
The most similar types are GANs that can control the output such as the conditional GAN \cite{Mirza2014}. 
The main structural differences from such GANs are 
that the character class information is provided only to the generator's input, 
and that the sampling from the real data distribution is associated with the character class. 
The procedure intrinsically makes GlyphGAN learn 
the conditional distribution of the target image given the class information. 

Legibility: 
In the legibility evaluation, 
we showed the learning method employed in the GlyphGAN is effective in improving the legibility of the generated fonts. 
Compared with GlyphGAN, DCGAN-based and WGAN-Clipping-based learning 
led to the collapse of the generated fonts. 
In the results where the DCGAN was used as the learning framework shown in Fig. \ref{05_result_change_framework}(a), 
almost the same patterns were generated even if different character class vectors were given. 
This is because of the phenomenon called mode collapse in which the output is biased to a specific pattern. 
In Fig. \ref{05_result_change_framework}(b), 
although WGAN-Clipping-based method generated fonts more efficiently than the DCGAN-based method, 
there were only a few patterns that could be recognized as letters. 
One possible explanation is that the WGAN-Clipping-based method 
could not represent the complexity of the data manifold owing to its approximated learning. 

Style consistency: 
Even though GlyphGAN employs unsupervised training in terms of style information, 
the generated fonts have a consistent style for all of the characters. 
However, this property is guaranteed by having a sufficient number of styles in the training data. 
If we use a training dataset that includes a few styles, 
the generated font does not guarantee style consistency, as shown in Fig. \ref{05_result_only10} and Table \ref{Consistency}.
This is because a large number of training data are required to learn the manifold of font styles. 

Limitations: This study includes some limitations. 
First, the dataset used in the experiment contains only alphabet letters. 
Font sets from different languages such as Chinese and Japanese 
can contain a larger number of characters than the alphabet; 
thus, expansion of the character class vector is required. 
Second, the legibility is not perfect. 
Although GlyphGAN improved the legibility as shown in Table \ref{05_result_recognition},
there still exists a 10 \% gap in the recognition accuracy between the generated fonts and existing fonts. 
The increase in the number of training data is one of the solutions for filling this gap. 
Finally, explicit style control is not performed. 
It is not obvious what type of $\bm{z}^{\rm s}$ to use 
for the generation of a specific font style. 
In this study, we obtained a latent space composed by $\bm{z}^{\rm s}$. 
Clarification of the relationship between the font design and the latent space using another framework
will lead to explicit style control.

	\section{Conclusion}
	In this paper, we proposed GlyphGAN a style-consistent font generation based on generative adversarial networks (GANs). 
In GlyphGAN, the input vector for the generator network consists of 
a character class vector and style vector,  
thereby allowing font generation with style consistency. 
In the font generation experiment, 
we showed that the learning method employed in the proposed method improved the legibility of the generated fonts. 
The experimental results also showed that the generated font set
had diversity different from the training patterns.

In future work, we will review the GAN structure to improve the quality of the generated font. 
Since many derivatives of GANs are still proposed even today, 
a better structure that enables more realistic generation can be found. 
Analysis of internal representations including the latent space
will be conducted to understand the generation process. 
Generation of multiple characters will also be investigated.
Finally, we plan to use vector images that have contour control points instead of bitmap images. 
This trial will lead to more practical font design without the limitation of resolution.

	\section*{Acknowledgment}
    This work was partially supported by JSPS KAKENHI Grant Number JP17H06100. 
 }




\begin{thebibliography}{10}
\expandafter\ifx\csname url\endcsname\relax
  \def\url#1{\texttt{#1}}\fi
\expandafter\ifx\csname urlprefix\endcsname\relax\def\urlprefix{URL }\fi
\expandafter\ifx\csname href\endcsname\relax
  \def\href#1#2{#2} \def\path#1{#1}\fi

\bibitem{Atarsaikhan2017}
G.~Atarsaikhan, B.~K. Iwana, A.~Narusawa, K.~Yanai, S.~Uchida, Neural font
  style transfer, in: Proceedings of the 14th International Conference on
  Document Analysis and Recognition (ICDAR), Vol.~5, 2017, pp. 51--56.

\bibitem{Chang2017}
J.~Chang, Y.~Gu, Chinese typography transfer, arXiv preprint arXiv:1707.04904.

\bibitem{Kaonashi2017}
\href{https://kaonashi-tyc.github.io/2017/04/06/zi2zi.html}{zi2zi: Master
  {Chinese} calligraphy with conditional adversarial networks}.
\newline\urlprefix\url{https://kaonashi-tyc.github.io/2017/04/06/zi2zi.html}

\bibitem{Lyu2017}
P.~Lyu, X.~Bai, C.~Yao, Z.~Zhu, T.~Huang, W.~Liu, Auto-encoder guided {GAN} for
  {Chinese} calligraphy synthesis, in: Proceedings of the 14th International
  Conference on Document Analysis and Recognition (ICDAR), 2017, pp.
  1095--1100.

\bibitem{Bernhardsson2016blog}
E.~Bernhardsson,
  \href{https://erikbern.com/2016/01/21/analyzing-50k-fonts-using-deep-neural-networks.html}{Analyzing
  50k fonts using deep neural networks}.
\newline\urlprefix\url{https://erikbern.com/2016/01/21/analyzing-50k-fonts-using-deep-neural-networks.html}

\bibitem{Bernhardsson2016git}
E.~Bernhardsson, \href{https://github.com/erikbern/deep-fonts}{deep-fonts}.
\newline\urlprefix\url{https://github.com/erikbern/deep-fonts}

\bibitem{Goodfellow2014}
I.~Goodfellow, J.~Pouget-Abadie, M.~Mirza, B.~Xu, D.~Warde-Farley, S.~Ozair,
  A.~Courville, Y.~Bengio, Generative adversarial nets, in: Proceedings of the
  Advances in Neural Information Processing Systems (NIPS), 2014, pp.
  2672--2680.

\bibitem{Devroye1995}
L.~Devroye, M.~McDougall, Random fonts for the simulation of handwriting,
  Electronic Publishing 8~(4) (1995) 281--294.

\bibitem{Tenenbaum2000}
J.~B. Tenenbaum, W.~T. Freeman, Separating style and content with bilinear
  models, Neural Computation 12~(6) (2000) 1247--1283.

\bibitem{Suveeranont2009}
R.~Suveeranont, T.~Igarashi, Feature-preserving morphable model for automatic
  font generation, in: Proceedings of the ACM SIGGRAPH ASIA 2009 Sketches,
  2009, p.~7.

\bibitem{Lake2015}
B.~M. Lake, R.~Salakhutdinov, J.~B. Tenenbaum, Human-level concept learning
  through probabilistic program induction, Science 350~(6266) (2015)
  1332--1338.

\bibitem{Miyazaki2017}
T.~Miyazaki, T.~Tsuchiya, Y.~Sugaya, S.~Omachi, M.~Iwamura, S.~Uchida, K.~Kise,
  Automatic generation of typographic font from a small font subset, arXiv
  preprint arXiv:1701.05703.

\bibitem{Yang2017}
S.~Yang, J.~Liu, Z.~Lian, Z.~Guo, Awesome typography: Statistics-based text
  effects transfer, in: Proceedings of the IEEE Conference on Computer Vision
  and Pattern Recognition (CVPR), 2017, pp. 7464--7473.

\bibitem{Wada2006}
A.~Wada, M.~Hagiwara, Japanese font automatic creating system reflecting user's
  {Kansei}, in: Proceedings of the IEEE International Conference on System, Man
  an Cybernetics, Vol.~4, 2003, pp. 3804--3809.

\bibitem{Wang2008}
Y.~Wang, H.~Wang, C.~Pan, L.~Fang, Style preserving {Chinese} character
  synthesis based on hierarchical representation of character, in: Proceedings
  of the IEEE International Conference on Acoustics, Speech and Signal
  Processing (ICASSP), 2008, pp. 1097--1100.

\bibitem{Campbell2014}
N.~D. Campbell, J.~Kautz, Learning a manifold of fonts, ACM Transactions on
  Graphics (TOG) 33~(4) (2014) 91.

\bibitem{Uchida2015}
S.~Uchida, Y.~Egashira, K.~Sato, Exploring the world of fonts for discovering
  the most standard fonts and the missing fonts, in: Proceedings of the 13th
  International Conference on Document Analysis and Recognition (ICDAR), 2015,
  pp. 441--445.

\bibitem{Gatys2016}
L.~A. Gatys, A.~S. Ecker, M.~Bethge, Image style transfer using convolutional
  neural networks, in: Proceedings of the IEEE Conference on Computer Vision
  and Pattern Recognition (CVPR), 2016, pp. 2414--2423.

\bibitem{Baluja2017}
S.~Baluja, Learning typographic style: from discrimination to synthesis,
  Machine Vision and Applications 28~(5-6) (2017) 551--568.

\bibitem{Isola2016}
P.~Isola, J.-Y. Zhu, T.~Zhou, A.~A. Efros, Image-to-image translation with
  conditional adversarial networks, in: Proceedings of the IEEE Conference on
  Computer Vision and Pattern Recognition (CVPR), 2017, pp. 1125--1134.

\bibitem{Odena2017}
A.~Odena, C.~Olah, J.~Shlens, Conditional image synthesis with auxiliary
  classifier {GANs}, in: Proceedings of the 34th International Conference on
  Machine Learning, 2017, pp. 2642--2651.

\bibitem{Taigman2017}
Y.~Taigman, A.~Polyak, L.~Wolf, Unsupervised cross-domain image generation,
  arXiv preprint arXiv:1611.02200.

\bibitem{Azadi2018}
S.~Azadi, M.~Fisher, V.~Kim, Z.~Wang, E.~Shechtman, T.~Darrell, Multi-content
  {GAN} for few-shot font style transfer, in: Proceedings of the IEEE
  Conference on Computer Vision and Pattern Recognition (CVPR), 2018, pp.
  7564--7573.

\bibitem{Lin2018}
X.~Lin, J.~Li, H.~Zeng, R.~Ji, Font generation based on least squares
  conditional generative adversarial nets, Multimedia Tools and Applications
  (2018) 1--15.

\bibitem{Guo2018}
Y.~Guo, Z.~Lian, Y.~Tang, J.~Xiao, Creating new {Chinese} fonts based on
  manifold learning and adversarial networks, in: O.~Diamanti, A.~Vaxman
  (Eds.), Proceedings of the Eurographics - Short Papers, The Eurographics
  Association, 2018.

\bibitem{Bhunia2018}
A.~K. Bhunia, A.~K. Bhunia, P.~Banerjee, A.~Konwer, A.~Bhowmick, P.~P. Roy,
  U.~Pal, Word level font-to-font image translation using convolutional
  recurrent generative adversarial networks, arXiv preprint arXiv:1801.07156.

\bibitem{Radford2016}
A.~Radford, L.~Metz, S.~Chintala, Unsupervised representation learning with
  deep convolutional generative adversarial networks, arXiv preprint
  arXiv:1511.06434.

\bibitem{Arjovsky2017}
M.~Arjovsky, S.~Chintala, L.~Bottou, Wasserstein generative adversarial
  networks, in: Proceedings of the 34th International Conference on Machine
  Learning, 2017, pp. 214--223.

\bibitem{Gulrajani2017}
I.~Gulrajani, F.~Ahmed, M.~Arjovsky, V.~Dumoulin, A.~C. Courville, Improved
  training of {Wasserstein} {GANs}, in: Proceedings of the Advances in Neural
  Information Processing Systems (NIPS), 2017, pp. 5769--5779.

\bibitem{Mirza2014}
M.~Mirza, S.~Osindero, Conditional generative adversarial nets, arXiv preprint
  arXiv:1411.1784.

\bibitem{Chen2016}
X.~Chen, Y.~Duan, R.~Houthooft, J.~Schulman, I.~Sutskever, P.~Abbeel,
  {InfoGAN}: Interpretable representation learning by information maximizing
  generative adversarial nets, in: Proceedings of the Advances in Neural
  Information Processing Systems (NIPS), 2016, pp. 2172--2180.

\bibitem{StarGAN2018}
Y.~Choi, M.~Choi, M.~Kim, J.-W. Ha, S.~Kim, J.~Choo, {StarGAN}: Unified
  generative adversarial networks for multi-domain image-to-image translation,
  in: Proceedings of the IEEE Conference on Computer Vision and Pattern
  Recognition (CVPR), 2018, pp. 8789--8797.

\bibitem{wang2018high}
T.-C. Wang, M.-Y. Liu, J.-Y. Zhu, A.~Tao, J.~Kautz, B.~Catanzaro,
  High-resolution image synthesis and semantic manipulation with conditional
  {GANs}, in: Proceedings of the IEEE Conference on Computer Vision and Pattern
  Recognition (CVPR), 2018, pp. 8798--8807.

\bibitem{shen2018faceid}
Y.~Shen, P.~Luo, J.~Yan, X.~Wang, X.~Tang, {FaceID-GAN}: Learning a symmetry
  three-player {GAN} for identity-preserving face synthesis, in: Proceedings of
  the IEEE Conference on Computer Vision and Pattern Recognition (CVPR), 2018,
  pp. 821--830.

\bibitem{Liang_2018_ECCV}
X.~Liang, H.~Zhang, L.~Lin, E.~Xing, Generative semantic manipulation with
  mask-contrasting {GAN}, in: Proceedings of the European Conference on
  Computer Vision (ECCV), 2018, pp. 574--590.

\bibitem{bodla2018semi}
N.~Bodla, G.~Hua, R.~Chellappa, Semi-supervised {FusedGAN} for conditional
  image generation, in: Proceedings of the European Conference on Computer
  Vision (ECCV), 2018, pp. 669--683.

\bibitem{Liu2016}
M.-Y. Liu, O.~Tuzel, Coupled generative adversarial networks, in: Proceedings
  of the Advances in Neural Information Processing Systems (NIPS), 2016, pp.
  469--477.

\bibitem{Mao2017}
X.~Mao, Q.~Li, H.~Xie, {AlignGAN}: Learning to align cross-domain images with
  conditional generative adversarial networks, arXiv preprint arXiv:1707.01400.

\bibitem{Kingma2015}
D.~P. Kingma, J.~Ba, Adam: A method for stochastic optimization, arXiv preprint
  arXiv:1412.6980.

\end{thebibliography}






\end{document}